\documentclass[letterpaper,10pt,conference]{ieeeconf}
\IEEEoverridecommandlockouts
\usepackage{times}
\usepackage{mathptmx}
\usepackage{amsmath,amssymb}
\usepackage{graphicx}
\usepackage{booktabs}
\usepackage{multirow}
\usepackage{array}
\usepackage{tabularx}
\usepackage{stfloats}
\usepackage{float}
\usepackage{balance}
\usepackage{url}

\newcommand{\scorearb}{ScoreArb}
\newcommand{\actPersist}{\textsc{Persist}}
\newcommand{\actSwitch}{\textsc{Switch}}
\newcommand{\actAbort}{\textsc{Abort}}
\newcommand{\actCommit}{\textsc{Commit}}
\newtheorem{proposition}{Proposition}
\newcounter{batnavalg}

\title{\LARGE \bf
BAT-Nav: Belief-Based Arbitration and Termination via Remaining Discoverability
in Multi-Goal Semantic Navigation
}

\author{
Xi~Lin$^{1,3}$,
Kangyi~Wu$^{2,3}$,
Jiayi~Li$^{3}$,
Jiaqiao~Tang$^{3}$,
Qingrong~He$^{3}$,
and Lin~Zhao$^{3,\dag}$
\thanks{$^{1}$LCSR, Johns Hopkins University; $^{2}$Xi'an Jiaotong University;
$^{3}$JD Explore Academy, Beijing, China. $^{\dag}$Corresponding author:
\texttt{zhaolins@foxmail.com}.}
}

\begin{document}
\maketitle
\thispagestyle{empty}
\pagestyle{empty}

\begin{abstract}
Multi-goal semantic navigation couples searches through a shrinking horizon:
effort spent on one request can make another effectively undiscoverable.
Discoverability therefore depends on both goal evidence and the budget that
other requests consume.

We present BAT-Nav, \emph{Belief-Based Arbitration and Termination}, which
estimates \emph{remaining discoverability}: the probability that a frozen
executor can complete a goal within an additional budget. A calibrated local
hazard converts navigation telemetry into a budget-conditioned completion
curve. Its marginal return governs reversible \actPersist{}/\actSwitch{};
conservative belief and local evidence govern \actAbort{}/\actCommit{}.

On intervention-independent replays, the model obtains Brier $.132$ and ECE
$.034$ without transfer refitting; 50-action completion rises from $.047$ to
$.321$ across marginal-return sextiles. On HM3D/ApexNav, BAT-Nav raises CR
from $.372$ to $.414$ and obtains MGSR $.173$. Its gain widens from $3.4$ to
$5.8$ CR points as competing goals increase from two to five, with the same
ordering on MP3D and under infeasible requests.
\end{abstract}

\section{INTRODUCTION}

Open-vocabulary navigation increasingly answers where to search for one active
target. VLFM~\cite{yokoyama2024vlfm}, ApexNav~\cite{zhang2025apexnav}, and
related systems combine geometric exploration with vision-language cues.

\begin{figure}[H]
\centering
\includegraphics[width=\columnwidth]{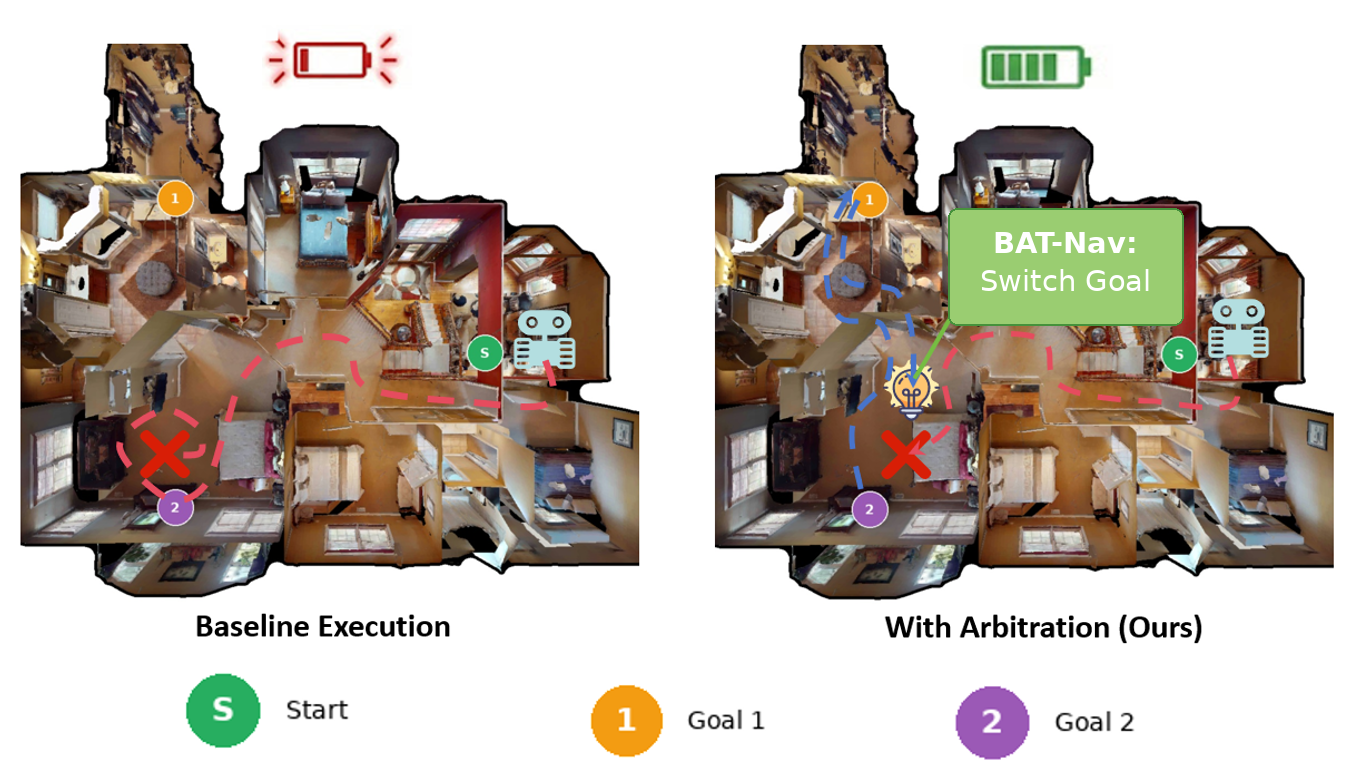}
\caption{\textbf{Shared-budget coupling.} A low-yield request consumes the
finite horizon while later unresolved requests wait.}
\label{fig:intro}
\end{figure}

A multi-goal mission must also allocate a shared time, battery, or action
budget. Every action assigned to one request shortens the others' horizon, so
a present target may become unrecoverable before it is attempted
(Fig.~\ref{fig:intro}).

We define \emph{remaining discoverability} as the probability that the current
executor can complete an unresolved goal within the remaining shared horizon.
Unlike existence, it decreases when viewpoints are weak, travel is costly, or
other goals consume the budget. Prior multi-object work emphasizes order and
memory~\cite{wani2020multion,gireesh2023sam,busch2025onemap}; remaining
discoverability asks whether continued search still justifies its cost.

This quantity introduces a state dimension absent from isolated single-goal
search. With $K=1$, the remaining-horizon argument of a goal changes only as
that search consumes actions. With several unresolved requests, it also
changes while other goals are active. Multi-goal navigation is therefore not
single-goal navigation repeated $K$ times: an untouched request can
deteriorate because execution is consumed elsewhere.
Section~\ref{sec:scaling} measures how this dimension strengthens as $K$
grows.

BAT-Nav builds arbitration around $F_g(b\mid h_t)$, the probability that goal
$g$ is completed if a total of $b$ additional primitive actions is devoted to
it from history $h_t$. The allocation need not be contiguous. Evaluating the
curve at the remaining horizon gives remaining discoverability, while its
marginal slope gives expected search return. The same object therefore
supports both \emph{which goal should receive effort next} and \emph{whether a
goal is still worth retaining}.

\begin{figure*}[t]
\centering
\includegraphics[width=0.86\textwidth]{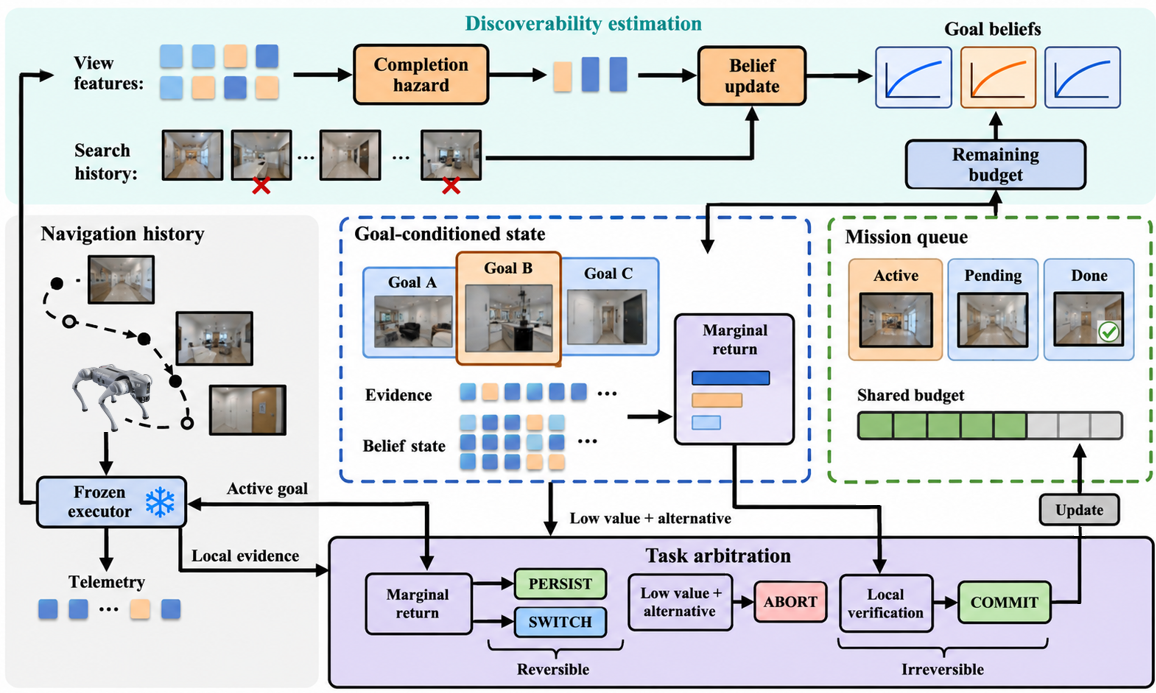}
\caption{\textbf{BAT-Nav system overview.} The navigation executor performs
mapping, exploration, waypoint generation, and low-level control. BAT-Nav
maintains goal-conditioned completion beliefs and updates the mission queue;
the same interface is used with VLFM and ApexNav.}
\label{fig:architecture}
\end{figure*}

\actPersist{} and \actSwitch{} are reversible because suspended goals can be
revisited; \actAbort{} and \actCommit{} are not. BAT-Nav therefore uses
point-estimate marginal return for reversible allocation and conservative
criteria for irreversible transitions. Figure~\ref{fig:architecture}
summarizes the interface above a frozen open-vocabulary executor.

Our contributions are threefold. \textbf{First}, we show that multi-goal
semantic navigation carries a state dimension that single-goal search does
not have: remaining discoverability changes not only with observations of a
request, but with how much of the shared horizon other requests consume; we
make this quantity explicit and estimable. \textbf{Second}, we estimate it
from navigation telemetry without interpreting backbone semantic scores as
probabilities, use its marginal return for allocation and permanent removal,
and keep commitment in a separate local sequential verifier. \textbf{Third},
we validate the state itself: the belief is calibrated against held-out
outcomes, the allocation index predicts future search return, and the benefit
grows as more goals compete for the same horizon.

\section{RELATED WORK}

\subsection{Open-Vocabulary Object Navigation}
VLFM~\cite{yokoyama2024vlfm} constructs language-conditioned frontier values;
ApexNav~\cite{zhang2025apexnav} adapts semantic and geometric exploration;
BeliefMapNav~\cite{zhou2025beliefmapnav} maintains a 3D target belief. These
methods improve search for one active target. BAT-Nav operates above the
executor and allocates a finite mission horizon among unresolved requests.
Goal-oriented semantic mapping, potential functions, predictive target priors,
and generative maps provide complementary map-based foundations
~\cite{chaplot2020semexp,ramakrishnan2022poni,zhai2023peanut,
zhang2024imagine}. Auxiliary objectives learn complementary exploration behavior
~\cite{ye2021auxiliary}. Recent
zero-shot systems inject commonsense or scene-graph structure into frontier
selection~\cite{zhou2023esc,yu2023l3mvn,yin2024sgnav,yin2025unigoal}, but
still optimize the active request rather than a shared queue. Their perception
stacks commonly build on language-image pretraining, open-set detection, and
promptable masks~\cite{radford2021clip,liu2024groundingdino,kirillov2023sam}.

\subsection{Multi-Object and Lifelong Navigation}
MultiON~\cite{wani2020multion} established ordered multi-object navigation.
Sequence-Agnostic Multi-Object Navigation~\cite{gireesh2023sam} receives all
requested classes simultaneously but primarily optimizes their visiting
order; BAT-Nav additionally decides whether a request should be suspended or
permanently removed when its expected return no longer justifies the shared
horizon it would consume. OneMap~\cite{busch2025onemap} reuses a persistent
open-vocabulary map across object queries, MOPA~\cite{raychaudhuri2024mopa}
combines modular perception and navigation for MultiON-style tasks, and
GOAT-Bench~\cite{khanna2024goat} evaluates long sequences of open-vocabulary
goals. Long-horizon VLN also exposes progress and memory drift; Dual-Anchoring
explicitly stabilizes both internal states~\cite{wu2026dualan}. BAT-Nav instead
tracks cross-request budget loss and makes reversible and irreversible queue
decisions above a frozen executor.

\subsection{Execution Monitoring and Resource Allocation}
Execution monitoring separates nominal control from continuation, recovery,
and termination~\cite{pettersson2005execution}; bandit methods
~\cite{auer2002ucb} allocate trials among uncertain alternatives. BAT-Nav
instead estimates a budget-conditioned completion curve whose value and slope
correspond to mission recoverability and expected search return.
This budgeted switching view is also related to temporally extended options
in semi-Markov decision processes~\cite{sutton1999options}.

\section{PROBLEM FORMULATION}

A mission contains semantic requests $\mathcal G=\{g_1,\ldots,g_K\}$ and a
global action budget $B_{\max}$. After $C_t$ primitive actions, the remaining
budget is
\begin{equation}
B_{\mathrm{rem}}(t)=B_{\max}-C_t.
\end{equation}
The frozen navigation executor receives the active request $g_t$ and
observation $o_t$ and produces
\begin{equation}
a_t\sim\pi_{\mathrm{nav}}(o_t,g_t).
\end{equation}
The mission objective is
\begin{equation}
\max_{\pi_{\mathrm{mission}}}
\mathbb E\left[\sum_{g\in\mathcal G}\mathbf 1[\mathrm{success}(g)]\right],
\qquad C_T\le B_{\max}.
\label{eq:mission}
\end{equation}

For unresolved goal $g$, define
\begin{equation}
F_g(b\mid h_t)=P\!\left(
\substack{g\text{ is completed if a total of }b\text{ additional}\\
\text{primitive actions is allocated to }g\text{ from }h_t}
\right).
\label{eq:F}
\end{equation}
The $b$ actions need not be contiguous: a request may be suspended and later
reactivated, and $b$ counts the total future execution assigned to that goal.
Remaining discoverability is
\begin{equation}
b_g(t)=F_g(B_{\mathrm{rem}}(t)\mid h_t).
\label{eq:belief}
\end{equation}
Equivalently, let $Z_g^t$ indicate that the frozen executor can still complete
$g$ within the remaining horizon. Then $b_g(t)=P(Z_g^t=1\mid h_t)$. This is
not physical existence: a present target can become undiscoverable because
reachable evidence is weak, topology is unfavorable, or too little mission
budget remains.

The definition creates the shared-budget coupling directly. If observations
relevant to $g$ are unchanged while another request consumes $\Delta B$
actions, monotonicity in the budget argument gives
\begin{equation}
F_g(B_{\mathrm{rem}}-\Delta B\mid h_t)
\le F_g(B_{\mathrm{rem}}\mid h_t).
\end{equation}
Thus an untouched request can lose discoverability purely because execution
was allocated elsewhere.

A relaxed future allocation is
\begin{equation}
\max_{u_g\ge0}\sum_{g\in\mathcal G_{\mathrm{unres}}}F_g(u_g\mid h_t),
\qquad
\mathrm{s.t.}\quad \sum_g u_g\le B_{\mathrm{rem}}(t).
\label{eq:alloc}
\end{equation}
BAT-Nav approximates this resource allocation online while treating permanent
removal and target commitment as separate irreversible decisions.

\section{BELIEF-BASED ARBITRATION AND TERMINATION}

\subsection{Completion Hazard and Discoverability}

We do not interpret the executor's semantic score as a probability. Instead,
we estimate a calibrated local completion hazard from geometry, coverage,
search history, and semantic cues, and update discoverability from unsuccessful
inspections.

For candidate viewpoint $v$, define
\begin{equation}
d_g(v)=P\!\left(C_g^{(N_d)}=1\mid v,\ g\ \mathrm{reachable},\ h_t\right).
\label{eq:hazard}
\end{equation}
Here, $C_g^{(N_d)}$ indicates completion within $N_d$ actions after inspecting $v$.
A logistic model fitted on development logs estimates
\begin{equation}
d_g(v)=\sigma(\theta^\top x_g(v)),
\end{equation}
where
\begin{equation}
x_g(v)=[\Delta A(v),r(v),\cos\phi(v),s_g(v),q_g^{\mathrm{cov}}(v),
\bar c_g(t),1]^\top.
\end{equation}
$\Delta A(v)$ is newly observable traversable area. $r(v)$ is the map
distance from $v$ to its associated goal-conditioned frontier or semantic
proposal, and $\phi(v)$ is the planned camera-bearing offset toward that
proposal; neither uses ground-truth target geometry. $s_g(v)$ is the
executor semantic search score, rank-normalized within the current candidate
set. $q_g^{\mathrm{cov}}(v)$ measures how much goal-conditioned frontier
quality has already been covered, and
$\bar c_g(t)=\min(c_g/(B_{\mathrm{rem}}+\epsilon),1)$ represents search
effort already spent on the goal. A training label is positive when the frozen executor completes the goal
within $N_d=15$ actions after visiting $v$. The semantic score is therefore one feature among geometric,
coverage, and search-history cues rather than a probability itself.

$\pi_0^g$ is the Beta$(1,1)$-smoothed frequency with which the frozen
executor completes category $g$ across development scenes. Unsuccessful
inspections at $v_1,\ldots,v_n$ update
\begin{equation}
\pi_t^g=
\frac{\pi_0^g\prod_i(1-d_g(v_i))}
{\pi_0^g\prod_i(1-d_g(v_i))+(1-\pi_0^g)}.
\label{eq:posterior}
\end{equation}
Candidate views are non-max suppressed so consecutive updates are separated
by at least $0.5$\,m or $30^\circ$. Spatial suppression keeps successive posterior updates approximately
independent.

Let $c_{\mathrm{search}}(v)$ be the primitive-action cost of the shortest
collision-free map path to $v$, computed from path length and heading change
using the executor's nominal forward and turn steps. Sort candidate viewpoints
by decreasing $d_g(v)/c_{\mathrm{search}}(v)$ and let $\mathcal V_g(u)$ be
the longest prefix whose cumulative cost does not exceed $u$. We construct
\begin{equation}
F_g(u\mid h_t)=
\pi_t^g\left[1-\prod_{v\in\mathcal V_g(u)}(1-d_g(v))\right],
\label{eq:Fconstruct}
\end{equation}
with piecewise-linear interpolation between cumulative viewpoint costs.
$F_g$ is non-decreasing, and its marginal gain per additional action decreases
along the sorted sequence, yielding a concave interpolated curve. Candidate
costs and hazards are recomputed after every allocation epoch.

\subsection{Marginal Search Return}

\begin{proposition}
If every $F_g(u\mid h_t)$ is non-decreasing and concave, the relaxed problem
in Eq.~\eqref{eq:alloc} equalizes active marginal returns among goals that
receive budget, up to the granularity of the candidate-viewpoint
discretization.
\end{proposition}
This is the standard KKT condition for separable concave resource allocation
~\cite{boyd2004convex}. BAT-Nav uses its greedy online analogue. For current
goal $g$ and candidate next goal $h$,
\begin{equation}
I_{g\rightarrow h}(t)=
\max_{v\in\mathcal V_h^t}
\frac{\pi_t^h d_h(v)}
{c_{\mathrm{search}}(v)+c_{\mathrm{switch}}(g,h)}.
\label{eq:index}
\end{equation}
The switch cost is zero for $h=g$. For $h\neq g$, it is the development-log
mean \emph{excess} relocation overhead after subtracting shortest-path travel
already included in $c_{\mathrm{search}}(v)$, conditioned on map distance to
$h$'s first candidate viewpoint. The index therefore avoids double-counting
travel and has units of expected completion per primitive action. If
$h^\star=\arg\max_h I_{g\rightarrow h}$ equals the active goal, BAT-Nav
\actPersist{}s; otherwise it \actSwitch{}es. A 20-action grace period and
the switch cost provide hysteresis. No separate anti-cycle rule is used:
revisiting a target without completion contributes further negative evidence
and lowers its future index.

\subsection{Conservative Irreversible Decisions}

\actAbort{} permanently removes the active request only when
\begin{equation}
F_g(B_{\mathrm{rem}}\mid h_t)<\varepsilon_{\min}
\quad\land\quad
\exists g'\neq g:\ I_{g\rightarrow g'}>I_{g\rightarrow g},
\label{eq:abort}
\end{equation}
with $\varepsilon_{\min}=0.05$. The first condition asks whether even the
full remaining horizon offers negligible completion probability; the second
retains the best available goal when all unresolved requests are difficult.

Commitment is a local encounter decision. Let $H_1$ denote evidence from a
valid target encounter and $H_0$ background or a distractor. Development
labels fit empirical densities $p_1(e)$ and $p_0(e)$. At sufficiently
separated viewpoints we accumulate
\begin{equation}
\Lambda_g=\sum_i
\mathrm{clip}\!\left(\log\frac{p_1(e_i)}{p_0(e_i)},
-\ell_{\max},\ell_{\max}\right),
\end{equation}
with $\ell_{\max}=2.0$. An SPRT-inspired verifier
~\cite{wald1945sequential} commits when
\begin{equation}
\Lambda_g>\log\frac{1-\beta_e}{\alpha_e},
\qquad \alpha_e=.05,\ \beta_e=.10.
\label{eq:commit}
\end{equation}
The densities are fitted offline; evaluation uses no privileged label.

\subsection{Online Controller and Configuration}

Algorithm~\ref{alg:batnav} evaluates irreversible decisions before reversible
allocation: \actCommit{}, then \actAbort{}, then \actSwitch{}, otherwise
\actPersist{}. Suspended goals retain their visited-viewpoint set and
posterior; reactivation does not reset $\pi_t^g$.

\refstepcounter{batnavalg}
\label{alg:batnav}
\begin{figure}[t]
\centering
\fbox{\begin{minipage}{0.94\columnwidth}
{\footnotesize\textbf{Algorithm~\thebatnavalg. BAT-Nav online arbitration.}}\\[-0.5mm]
{\scriptsize
1. Execute the frozen active-goal navigator and update candidate views,
$d_g$, $\pi_t^g$, and $\Lambda_g$.\\
2. If Eq.~\eqref{eq:commit} holds, move $g_a$ to completed and activate the
highest-index unresolved goal.\\
3. Else, after the grace period, if Eq.~\eqref{eq:abort} holds, move $g_a$ to
aborted and activate the highest-index unresolved goal.\\
4. Else set $h^\star=\arg\max_h I_{g_a\rightarrow h}$; switch if
$h^\star\neq g_a$, otherwise persist.}
\end{minipage}}
\end{figure}

The method changes neither backbone weights nor the low-level policy. Its
learned quantities are lightweight development-log calibrations: completion-hazard coefficients, local evidence densities, and the switch-cost model.
Table~\ref{tab:config} lists the complete protocol and configuration.

\begin{table}[t]
\centering
\caption{\textbf{Evaluation protocol and BAT-Nav configuration.}}
\label{tab:config}
\scriptsize
\setlength{\tabcolsep}{2.0pt}
\renewcommand{\arraystretch}{0.98}
\begin{tabularx}{0.98\columnwidth}{@{}p{0.39\columnwidth}X@{}}
\toprule
Item & Setting\\
\midrule
HM3D main set & 16 dev / 64 held-out scenes; 5000 missions\\
MP3D transfer & 2000 missions; no refitting\\
Main $K$/budget & $K=2$: 500; $K=3$: 650 actions\\
Hazard calibration & Logistic regression; $N_d=15$; 0.5\,m/$30^\circ$ view suppression\\
Prior / switch cost & Category completion frequency, Beta$(1,1)$ smoothed; map-distance relocation model\\
Permanent removal & $\varepsilon_{\min}=0.05$\\
Local verification & $\alpha_e=.05$, $\beta_e=.10$, $\ell_{\max}=2.0$\\
Allocation period & 20 primitive actions\\
\bottomrule
\end{tabularx}
\end{table}

Episode generation uses seeds $\{13,17,23,29,31\}$, a 24-category semantic
pool, per-scene caps of 80 HM3D and 120 MP3D missions, and a 1.0\,m geodesic
success radius.

\section{EXPERIMENTS}

We evaluate in Habitat~\cite{savva2019habitat} using HM3D
~\cite{ramakrishnan2021hm3d} for development and held-out testing and MP3D
~\cite{chang2017mp3d} for cross-dataset transfer. We ask four questions.
\textbf{Q1} (Sec.~\ref{sec:mechanism}): is remaining discoverability
calibrated against held-out outcomes? \textbf{Q2}
(Sec.~\ref{sec:mechanism}): does the marginal-return index predict what the
executor will accomplish next? \textbf{Q3} (Sec.~\ref{sec:results}): does
belief-based arbitration improve multi-goal navigation across executors and
datasets? \textbf{Q4} (Secs.~\ref{sec:scaling} and \ref{sec:absent}): does
its advantage grow as budget competition or request infeasibility increases?

\subsection{Executors and Baselines}

VLFM~\cite{yokoyama2024vlfm} and ApexNav~\cite{zhang2025apexnav} are used as
alternative frozen active-goal executors. \textbf{Sequential} follows fixed
request order. \textbf{DynamicCap} assigns the active request
$\min(300,\max(B_{\mathrm{rem}}/N_t,50))$ actions, where $N_t$ is the
number of unresolved requests; it is included on VLFM/HM3D to isolate budget
capping. \textbf{UCB-Queue} treats unresolved requests as arms, reevaluating
every 20 actions with score
$\hat\mu_g+0.5\sqrt{\log(1+t)/(1+n_g)}$, where $\hat\mu_g$ is the
running mean of normalized progress and semantic evidence and $n_g$ is the
activation count. It never permanently removes an unresolved arm.
\textbf{PPO-Arbitrator} receives matched mission telemetry and learns the same
four queue actions with PPO~\cite{schulman2017ppo} from HM3D development
scenes. It is trained for two million environment steps over five seeds with
reward $+1$ per completed request, $-0.002$ per primitive action, $-0.20$ per
false commit, and $-0.10$ per false abort. PPO requires two million online
environment steps per executor, whereas the remaining controllers require
only offline log fitting; we therefore report it on VLFM only.
\textbf{ScoreArb} is our strongest non-belief arbitration baseline: fixed
scores combine progress, semantic evidence, retained budget, and temporal
stability. Its permanent-removal event fires after the same 20-action grace
period when its search-feasibility score falls below $0.30$. Both ScoreArb
and BAT-Nav use the same local sequential verifier in Eq.~\eqref{eq:commit}
for \actCommit{}; they differ only in how allocation and permanent-removal
state are represented.

CR is the fraction of requested goals completed; MGSR requires all requests
in a mission to complete. WSF is the fraction of executed actions attributed
to goals unresolved at mission end. Steps/Comp. is total actions divided by
total completed requests. Monopoly is the largest fraction of mission actions
consumed by one unresolved request, averaged over missions. Deterministic
controllers are compared with 10,000 paired bootstrap resamples over matched
missions; PPO values average five independently trained seeds.

\section{MECHANISM VALIDATION}
\label{sec:mechanism}

\subsection{Calibration on Intervention-Independent Replays}

Calibration cannot use BAT-Nav's own rollout labels because an aborted goal
would then be unsuccessful by construction. We therefore use only Sequential
rollouts. From 500 held-out missions we save states every 25 actions and
evaluate every fourth snapshot, yielding approximately $6.7$k unresolved
goal-state pairs. From each state the frozen executor is replayed with the
remaining mission budget; completion provides the evaluation label for
Eq.~\eqref{eq:belief}.

\begin{figure}[t]
\centering
\includegraphics[width=\columnwidth]{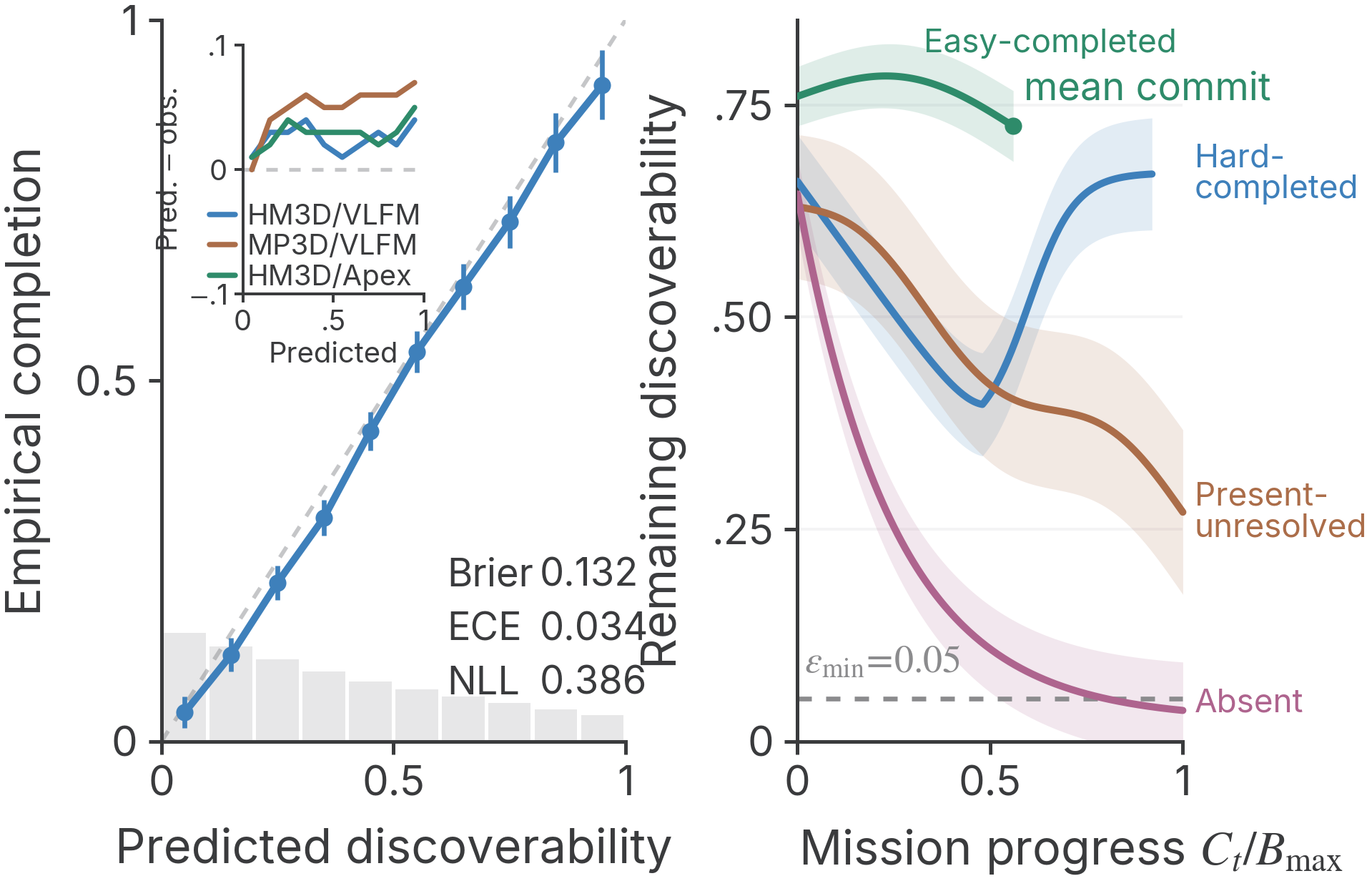}
\caption{\textbf{Remaining-discoverability validation.} (a) HM3D/VLFM
reliability with mission-clustered bootstrap 95\% intervals and prediction-density histogram;
inset: prediction-minus-observation residuals for transfer settings without
refitting. (b) Mean trajectories by final outcome; bands are mission-clustered bootstrap
95\% intervals and the dashed line marks $\varepsilon_{\min}=0.05$.}
\label{fig:belief}
\end{figure}

Geometry, coverage, and search-history features alone give hazard ROC-AUC
$.692$; adding the backbone semantic score raises it to $.748$. On held-out
HM3D/VLFM states, remaining discoverability obtains Brier $.132$, NLL $.386$,
and ECE $.034$. Without refitting, MP3D/VLFM gives $.145/.414/.049$ and
HM3D/ApexNav gives $.139/.401/.042$ for Brier/NLL/ECE.
Figure~\ref{fig:belief}(a) shows the primary HM3D/VLFM reliability curve,
while its residual inset exposes the small transfer deviations around the
diagonal.

Figure~\ref{fig:belief}(b) groups trajectories into four outcome classes.
Among completed requests, easy-completed and hard-completed are the lowest and
highest quartiles of action cost to completion; present-unresolved requests
have a valid scene instance but remain incomplete, and absent requests have no
valid instance. Easy goals remain high until commitment and are truncated at
the mean commit time. Absent-goal discoverability decreases toward
$\varepsilon_{\min}$; hard-completed goals can recover when a newly exposed
region creates high-hazard viewpoints; present-unresolved requests decay with
larger uncertainty. Mission progress is normalized by $C_t/B_{\max}$ so
$K=2$ and $K=3$ episodes are comparable.

\begin{table*}[!t]
\centering
\caption{\textbf{Main multi-goal navigation results.} Best result within each
executor block is bold. DynamicCap is included only on HM3D/VLFM to isolate
budget capping; PPO is reported only with VLFM because it requires separate
online training for each executor.}
\label{tab:main}
\scriptsize
\setlength{\tabcolsep}{2.6pt}
\renewcommand{\arraystretch}{1.0}
\begin{tabular*}{0.98\textwidth}{@{\extracolsep{\fill}}llccccc@{}}
\toprule
Dataset / Executor & Controller & CR$\uparrow$ & MGSR$\uparrow$ & WSF$\downarrow$ & Steps/Comp.$\downarrow$ & Mono.$\downarrow$\\
\midrule
\multirow{6}{*}{HM3D / VLFM}
& Sequential & .220 & .020 & .903 & 1039 & .39\\
& DynamicCap & .282 & .047 & .735 & 745 & .31\\
& UCB-Queue & .300 & .064 & .712 & 684 & .27\\
& PPO-Arbitrator & .307 & .069 & .704 & 661 & .26\\
& \scorearb & .345 & .104 & .682 & 555 & .18\\
& \textbf{BAT-Nav} & \textbf{.384} & \textbf{.138} & \textbf{.624} & \textbf{486} & \textbf{.12}\\
\midrule
\multirow{4}{*}{HM3D / ApexNav}
& Sequential & .276 & .050 & .810 & 803 & .32\\
& UCB-Queue & .332 & .088 & .700 & 619 & .24\\
& \scorearb & .372 & .132 & .625 & 508 & .15\\
& \textbf{BAT-Nav} & \textbf{.414} & \textbf{.173} & \textbf{.548} & \textbf{438} & \textbf{.08}\\
\midrule
\multirow{5}{*}{MP3D / VLFM}
& Sequential & .158 & .011 & .942 & 1462 & .44\\
& UCB-Queue & .218 & .034 & .790 & 963 & .36\\
& PPO-Arbitrator & .225 & .039 & .782 & 923 & .35\\
& \scorearb & .263 & .066 & .748 & 745 & .27\\
& \textbf{BAT-Nav} & \textbf{.296} & \textbf{.094} & \textbf{.688} & \textbf{640} & \textbf{.20}\\
\midrule
\multirow{4}{*}{MP3D / ApexNav}
& Sequential & .205 & .029 & .854 & 1104 & .41\\
& UCB-Queue & .250 & .056 & .760 & 849 & .31\\
& \scorearb & .292 & .087 & .694 & 679 & .23\\
& \textbf{BAT-Nav} & \textbf{.331} & \textbf{.121} & \textbf{.626} & \textbf{562} & \textbf{.16}\\
\bottomrule
\end{tabular*}
\end{table*}

\setcounter{figure}{4}
\begin{figure*}[!t]
\centering
\includegraphics[width=0.96\textwidth]{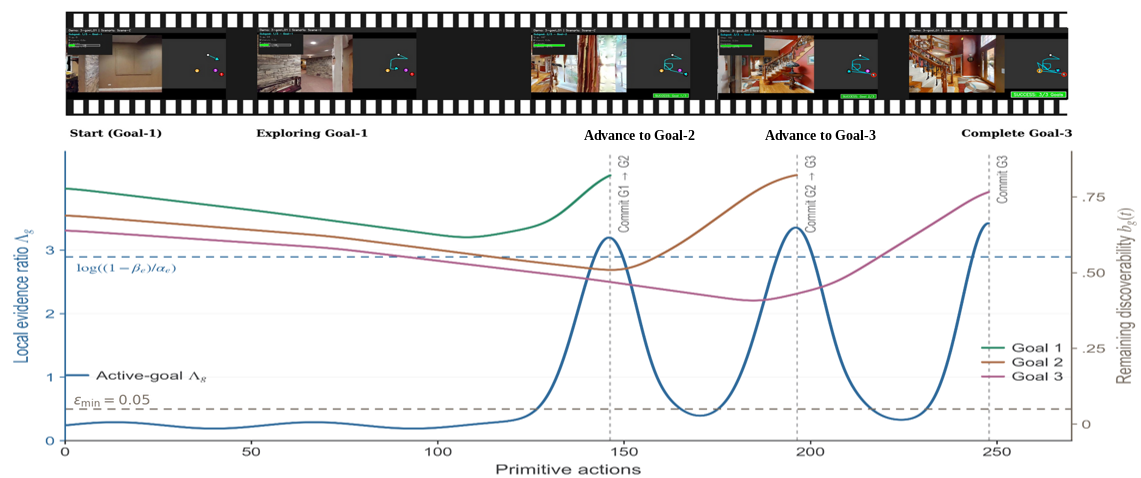}
\caption{\textbf{Synchronized simulation trace.} The filmstrip and mission
signals share the same primitive-action axis. The left axis shows the local
sequential verifier $\Lambda_g$; the right axis shows remaining
discoverability for the three requested goals. Commit and queue-advance
markers are aligned with the corresponding simulation frames.}
\label{fig:simulation}
\end{figure*}
\setcounter{figure}{3}

These checks matter because arbitration consumes probabilities across budgets,
not merely their ranking. A monotone but overconfident estimator may preserve
ROC-AUC while triggering premature permanent removals.

\subsection{Marginal Return Predicts Short-Horizon Completion}

\begin{figure}[H]
\centering
\includegraphics[width=\columnwidth]{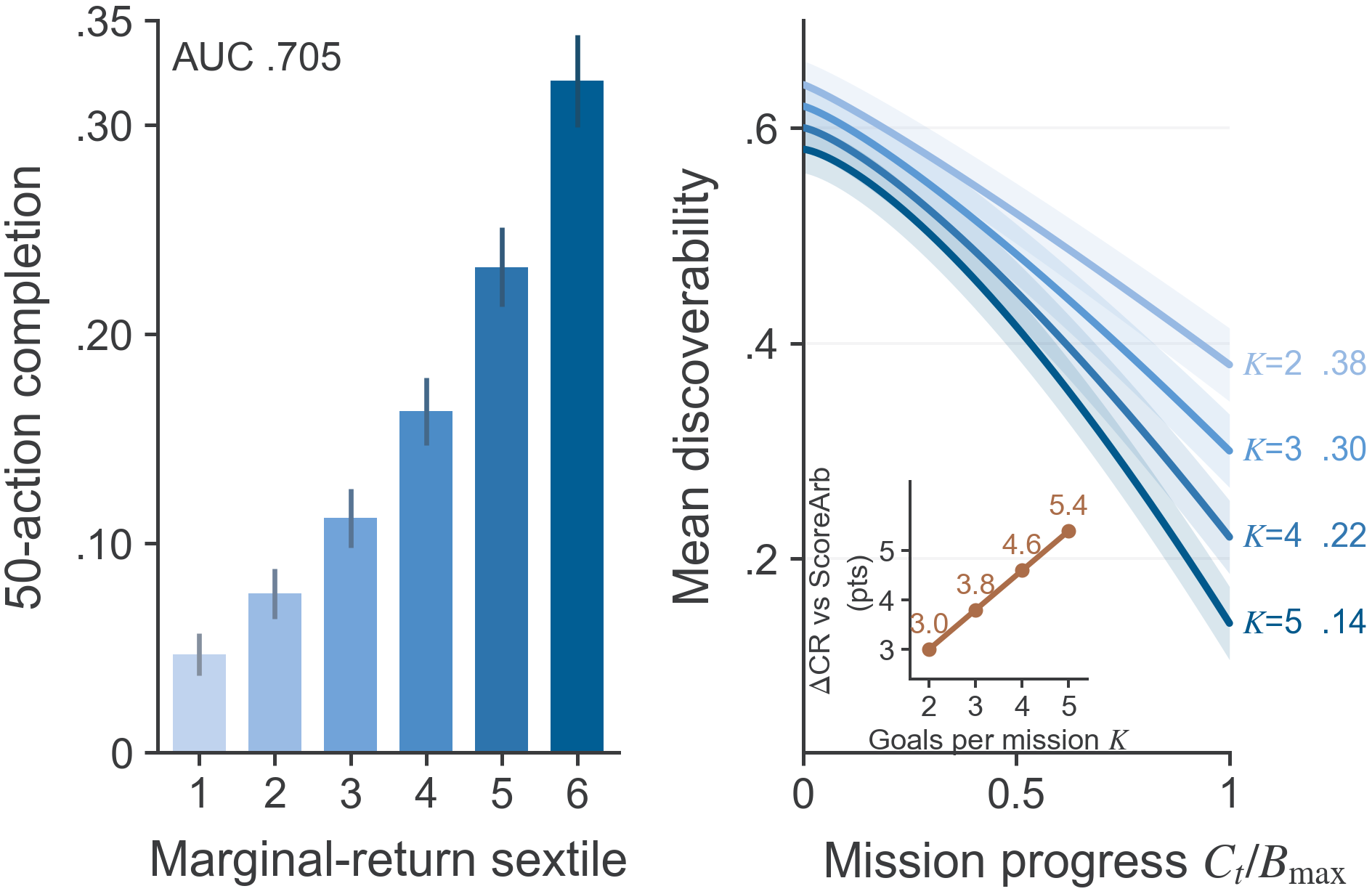}
\caption{\textbf{Marginal return and budget competition.} (a) Empirical
50-action completion across six equal-mass index sextiles; boundaries are
listed in Sec.~\ref{sec:mechanism}. (b) Average unresolved-goal
discoverability falls faster as $K$ grows; inset: BAT-Nav CR gain over
ScoreArb. Bands and error bars are mission-clustered bootstrap 95\% intervals.}
\label{fig:index}
\end{figure}

For index validation we independently subsample 350 Sequential missions and
replay approximately $4.9$k unresolved goal-state pairs for 50 actions. The
binary label is whether the replay completes the goal. Across six equal-mass
index sextiles, completion rises monotonically from $.047$, $.076$, $.112$,
$.163$, and $.232$ to $.321$; the index obtains ROC-AUC $.705$.
Figure~\ref{fig:index}(a) reports mission-clustered bootstrap intervals. The sextile boundaries
are $<.0005$, $.0005$--$.0011$, $.0011$--$.0019$, $.0019$--$.0030$,
$.0030$--$.0048$, and $\geq.0048$ expected completions per primitive action.
This directly tests the quantity used by the allocation policy rather than
only the final navigation score.

\setcounter{figure}{5}

\section{NAVIGATION RESULTS}
\label{sec:results}

Table~\ref{tab:main} reports matched comparisons under the same 500/650-action
horizon and 24-category pool. BAT-Nav and ScoreArb receive the same telemetry,
use the same local COMMIT verifier, and expose the same four queue actions;
they differ only in how allocation and permanent-removal state are
represented. Their difference is therefore concentrated in states where a
fixed score threshold and a calibrated budget-conditioned belief disagree.
Such states occupy only a minority of ordinary missions, but become more
frequent as budget competition or request infeasibility increases.

Relative to ScoreArb, BAT-Nav gains $+3.9$, $+4.2$, $+3.3$, and $+3.9$ CR
points on HM3D/VLFM, HM3D/ApexNav, MP3D/VLFM, and MP3D/ApexNav. Paired 95\%
bootstrap intervals are $[+.028,+.050]$, $[+.030,+.054]$,
$[+.022,+.044]$, and $[+.027,+.051]$; all exclude zero, and WSF,
Steps/Comp., and monopoly improve in every setting. Sections~\ref{sec:scaling}
and \ref{sec:absent} test the predicted higher-disagreement regimes.

\subsection{Scaling the Number of Competing Goals}
\label{sec:scaling}

We hold $B_{\max}=650$ fixed and vary $K\in\{2,3,4,5\}$ with ApexNav on
HM3D, isolating goal count while keeping the controller and horizon fixed.

\begin{table}[t]
\centering
\caption{\textbf{Goal-count scaling on HM3D/ApexNav with a fixed 650-action
horizon.} Cells report CR / MGSR.}
\label{tab:k}
\scriptsize
\setlength{\tabcolsep}{3.5pt}
\begin{tabular}{@{}ccccc@{}}
\toprule
$K$ & Sequential & UCB & \scorearb & \textbf{BAT-Nav}\\
\midrule
2 & .333/.084 & .369/.127 & .420/.178 & \textbf{.454/.214}\\
3 & .225/.032 & .294/.067 & .328/.094 & \textbf{.370/.125}\\
4 & .182/.009 & .237/.026 & .266/.041 & \textbf{.317/.068}\\
5 & .145/.002 & .188/.008 & .205/.016 & \textbf{.263/.035}\\
\bottomrule
\end{tabular}
\end{table}

The BAT-Nav minus ScoreArb margin grows from $3.4$ to $5.8$ CR points as
$K$ rises from 2 to 5, while Fig.~\ref{fig:index}(b) shows faster belief
decay. The widening gap therefore tracks stronger budget coupling.

\subsection{Infeasible Requests}
\label{sec:absent}

We replace 25\% of requested categories by scene-absent categories. Sequential
and UCB never remove requests; ScoreArb retains its fixed feasibility threshold,
whereas BAT-Nav uses Eq.~\eqref{eq:abort}.

\begin{table}[t]
\centering
\caption{\textbf{Infeasible-request stress test on HM3D/ApexNav at
$p_{\mathrm{abs}}=.25$.} Abort metrics apply only to controllers with
permanent removal.}
\label{tab:absent}
\scriptsize
\setlength{\tabcolsep}{1.8pt}
\begin{tabular}{@{}lcccccc@{}}
\toprule
Controller & CR$\uparrow$ & Feas.$\uparrow$ & WSF$\downarrow$ & P.$\uparrow$ & R.$\uparrow$ & Med.$\downarrow$\\
\midrule
Sequential & .205 & .273 & .872 & -- & -- & --\\
UCB-Queue & .236 & .315 & .780 & -- & -- & --\\
\scorearb & .270 & .360 & .690 & .76 & .68 & 102\\
BAT-Nav & \textbf{.308} & \textbf{.410} & \textbf{.612} & \textbf{.89} & \textbf{.83} & \textbf{68}\\
\bottomrule
\end{tabular}
\end{table}

BAT-Nav limits the feasible-goal CR decline to $.004$ versus ScoreArb's $.012$,
while improving abort precision/recall from $.76/.68$ to $.89/.83$ and median
removal time from 102 to 68 actions.

\subsection{Component Analysis}
On HM3D/ApexNav, removing semantic input, using fixed-horizon
discoverability, or setting uniform $\pi_0=.5$ changes CR/WSF from $.414/.548$
to $.401/.571$, $.395/.585$, and $.409/.557$, respectively. Across
$\pi_0\in\{.3,.5,.7,.9\}$, CR remains within $[.404,.414]$.

\section{PHYSICAL NAVIGATION}

Across 30 matched quadruped missions in three layouts, Sequential, ScoreArb,
and BAT-Nav obtain CR/MGSR of $.43/.13$, $.57/.23$, and $.60/.27$, with WSF
$.51$, $.35$, and $.31$. No HM3D model is refit. Five of six removed-object
runs end in removal (median 76 actions); arbitration takes $3.4$\,ms onboard.

\begin{figure}[t]
\centering
\includegraphics[width=\columnwidth]{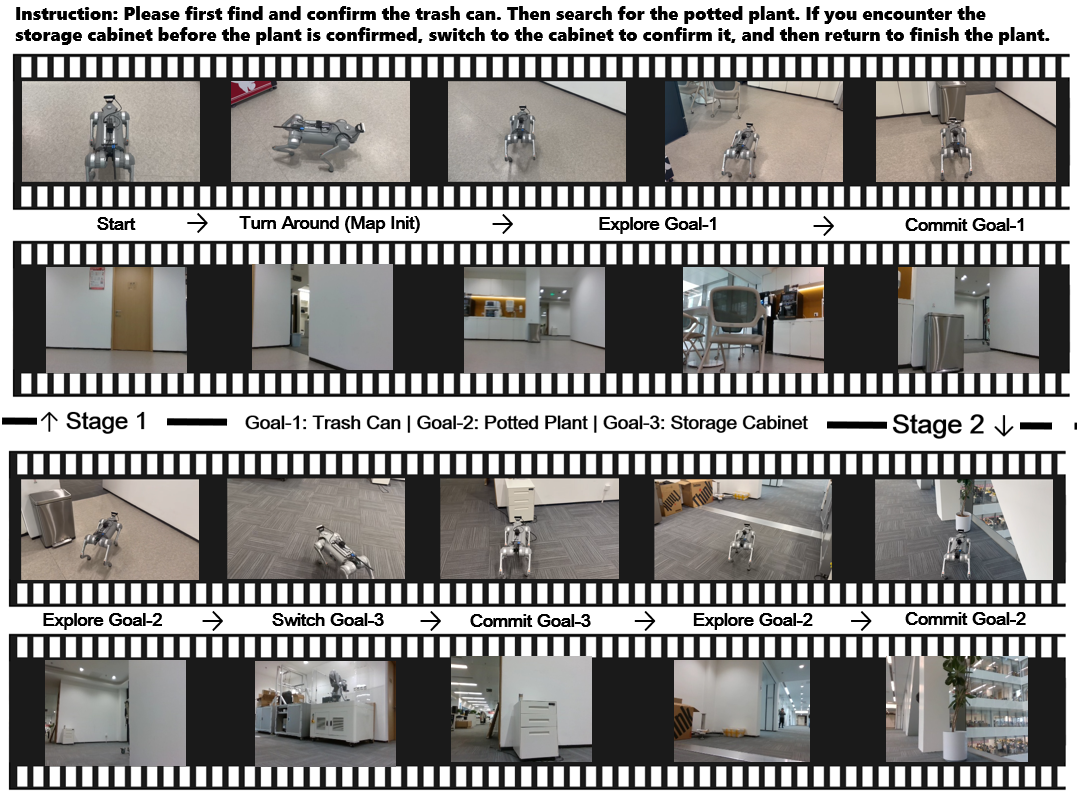}
\caption{\textbf{Representative physical multi-goal execution.} The robot
retains one mission queue across searches and returns to a suspended request
after completing another target. This visualized mission contains no absent
request; removed-object behavior is summarized in the text.}
\label{fig:physical}
\end{figure}

\section{DISCUSSION AND CONCLUSION}

The value of mission-level arbitration is not simply that it stops sooner,
but that it estimates what continued search costs. Four results support this
view. Remaining discoverability is calibrated against intervention-independent
future completion; the marginal-return index predicts short-horizon success;
its advantage grows as more goals compete for one horizon; and, under absent
requests, improved calibration translates directly into earlier and more
accurate permanent removal.

Executor capability and mission-level arbitration are complementary rather
than substitutable. A stronger executor raises the quality of each active
search, while arbitration determines how that capability is distributed across
requests. Their gains therefore compose. Under infeasible requests, BAT-Nav's
abort precision exceeds ScoreArb's by about 17\% relative, while feasible-goal
CR improves by about 14\%. Better mission-level judgement does not translate
one-for-one into mission success because the executor still bounds what any
reallocation can recover.

BAT-Nav therefore separates two coupled problems: the executor decides how to
search the active request, while the mission layer estimates whether unresolved
searches remain worth the shrinking horizon. Remaining discoverability supplies
a common state for reversible allocation and permanent removal, with local
evidence reserved for commitment. Large sensing shifts may require lightweight estimator recalibration, but no
navigation-policy retraining is needed.

\section*{ACKNOWLEDGMENT}
OpenAI ChatGPT assisted with language editing and with plotting code used to
render Figs.~3--5. The authors verified all technical content, experiments,
analyses, figures, and results.

\balance


\begingroup
\footnotesize
\begin{thebibliography}{99}
\setlength{\itemsep}{0pt}
\setlength{\parsep}{0pt}
\setlength{\parskip}{0pt}

\bibitem{yokoyama2024vlfm}
N. Yokoyama, S. Ha, D. Batra, J. Wang, and B. Bucher,
``VLFM: Vision-Language Frontier Maps for Zero-Shot Semantic Navigation,''
in \emph{Proc. IEEE Int. Conf. Robotics and Automation (ICRA)}, 2024.

\bibitem{zhang2025apexnav}
M. Zhang, Y. Du, C. Wu, J. Zhou, Z. Qi, J. Ma, and B. Zhou,
``ApexNav: An Adaptive Exploration Strategy for Zero-Shot Object Navigation
with Target-Centric Semantic Fusion,'' \emph{IEEE Robot. Autom. Lett.},
vol. 10, no. 11, pp. 11530--11537, 2025.

\bibitem{zhou2025beliefmapnav}
Z. Zhou, Y. Hu, L. Zhang, Z. Li, and S. Chen,
``BeliefMapNav: 3D Voxel-Based Belief Map for Zero-Shot Object Navigation,''
in \emph{Advances in Neural Information Processing Systems}, vol. 38, 2025.

\bibitem{wani2020multion}
S. Wani, S. Patel, U. Jain, A. X. Chang, and M. Savva,
``MultiON: Benchmarking Semantic Map Memory using Multi-Object Navigation,''
in \emph{Advances in Neural Information Processing Systems}, vol. 33,
pp. 9700--9712, 2020.

\bibitem{gireesh2023sam}
N. Gireesh, A. Agrawal, A. Datta, S. Banerjee, M. Sridharan,
B. Bhowmick, and M. Krishna,
``Sequence-Agnostic Multi-Object Navigation,'' in \emph{Proc. IEEE Int. Conf.
Robotics and Automation (ICRA)}, pp. 9573--9579, 2023.

\bibitem{busch2025onemap}
F. L. Busch, T. Homberger, J. Ortega-Peimbert, Q. Yang, and O. Andersson,
``One Map to Find Them All: Real-time Open-Vocabulary Mapping for Zero-shot
Multi-Object Navigation,'' in \emph{Proc. IEEE Int. Conf. Robotics and
Automation (ICRA)}, pp. 14835--14842, 2025.

\bibitem{raychaudhuri2024mopa}
S. Raychaudhuri, T. Campari, U. Jain, M. Savva, and A. X. Chang,
``MOPA: Modular Object Navigation With PointGoal Agents,'' in
\emph{Proc. IEEE/CVF Winter Conf. Applications of Computer Vision (WACV)},
pp. 5763--5773, 2024.

\bibitem{khanna2024goat}
M. Khanna \emph{et al.},
``GOAT-Bench: A Benchmark for Multi-Modal Lifelong Navigation,'' in
\emph{Proc. IEEE/CVF Conf. Computer Vision and Pattern Recognition (CVPR)},
pp. 16373--16383, 2024.

\bibitem{pettersson2005execution}
O. Pettersson,
``Execution Monitoring in Robotics: A Survey,'' \emph{Robotics and
Autonomous Systems}, vol. 53, no. 2, pp. 73--88, 2005.

\bibitem{auer2002ucb}
P. Auer, N. Cesa-Bianchi, and P. Fischer,
``Finite-Time Analysis of the Multiarmed Bandit Problem,''
\emph{Machine Learning}, vol. 47, pp. 235--256, 2002.

\bibitem{schulman2017ppo}
J. Schulman, F. Wolski, P. Dhariwal, A. Radford, and O. Klimov,
``Proximal Policy Optimization Algorithms,'' arXiv:1707.06347, 2017.

\bibitem{savva2019habitat}
M. Savva \emph{et al.},
``Habitat: A Platform for Embodied AI Research,'' in
\emph{Proc. IEEE Int. Conf. Computer Vision (ICCV)}, 2019.

\bibitem{ramakrishnan2021hm3d}
S. K. Ramakrishnan \emph{et al.},
``Habitat-Matterport 3D Dataset (HM3D): 1000 Large-Scale 3D Environments for
Embodied AI,'' in \emph{Proc. NeurIPS Datasets and Benchmarks}, 2021.

\bibitem{chang2017mp3d}
A. Chang \emph{et al.},
``Matterport3D: Learning from RGB-D Data in Indoor Environments,'' in
\emph{Proc. Int. Conf. 3D Vision (3DV)}, pp. 667--676, 2017.

\bibitem{boyd2004convex}
S. Boyd and L. Vandenberghe, \emph{Convex Optimization}.
Cambridge University Press, 2004.

\bibitem{wald1945sequential}
A. Wald, ``Sequential Tests of Statistical Hypotheses,''
\emph{Annals of Mathematical Statistics}, vol. 16, no. 2,
pp. 117--186, 1945.

\bibitem{chaplot2020semexp}
D. S. Chaplot, D. P. Gandhi, A. Gupta, and R. Salakhutdinov,
``Object Goal Navigation using Goal-Oriented Semantic Exploration,'' in
\emph{Advances in Neural Information Processing Systems}, vol. 33, 2020.

\bibitem{ramakrishnan2022poni}
S. K. Ramakrishnan, D. S. Chaplot, Z. Al-Halah, J. Malik, and K. Grauman,
``PONI: Potential Functions for ObjectGoal Navigation With Interaction-Free
Learning,'' in \emph{Proc. IEEE/CVF Conf. Computer Vision and Pattern
Recognition (CVPR)}, pp. 18890--18900, 2022.



\bibitem{zhai2023peanut}
A. J. Zhai and S. Wang, ``PEANUT: Predicting and Navigating to Unseen
Targets,'' in \emph{Proc. IEEE/CVF Int. Conf. Computer Vision (ICCV)},
pp. 10926--10935, 2023.

\bibitem{zhang2024imagine}
S. Zhang, X. Yu, X. Song, X. Wang, and S. Jiang, ``Imagine Before Go:
Self-Supervised Generative Map for Object Goal Navigation,'' in
\emph{Proc. IEEE/CVF Conf. Computer Vision and Pattern Recognition (CVPR)},
pp. 16414--16425, 2024.

\bibitem{ye2021auxiliary}
J. Ye, D. Batra, A. Das, and E. Wijmans, ``Auxiliary Tasks and Exploration
Enable ObjectGoal Navigation,'' in \emph{Proc. IEEE/CVF Int. Conf. Computer
Vision (ICCV)}, pp. 16117--16126, 2021.



\bibitem{zhou2023esc}
K. Zhou, K. Zheng, C. Pryor, Y. Shen, H. Jin, L. Getoor, and X. E. Wang,
``ESC: Exploration with Soft Commonsense Constraints for Zero-shot Object
Navigation,'' in \emph{Proc. Int. Conf. Machine Learning (ICML)},
pp. 42829--42842, 2023.

\bibitem{yu2023l3mvn}
B. Yu, H. Kasaei, and M. Cao, ``L3MVN: Leveraging Large Language Models for
Visual Target Navigation,'' in \emph{Proc. IEEE/RSJ Int. Conf. Intelligent
Robots and Systems (IROS)}, pp. 3554--3560, 2023.

\bibitem{yin2024sgnav}
H. Yin, X. Xu, Z. Wu, J. Zhou, and J. Lu, ``SG-Nav: Online 3D Scene Graph
Prompting for LLM-Based Zero-Shot Object Navigation,'' in \emph{Advances in
Neural Information Processing Systems}, vol. 37, pp. 5285--5307, 2024.

\bibitem{yin2025unigoal}
H. Yin, X. Xu, L. Zhao, Z. Wang, J. Zhou, and J. Lu, ``UniGoal: Towards
Universal Zero-shot Goal-oriented Navigation,'' in \emph{Proc. IEEE/CVF Conf.
Computer Vision and Pattern Recognition (CVPR)}, pp. 19057--19066, 2025.

\bibitem{wu2026dualan}
K. Wu, P. Li, K. Lyu, X. Lin, L. Zhao, Q. He, J. Wang, and J. Liu,
``Dual-Anchoring: Addressing State Drift in Vision-Language Navigation,'' in
\emph{Proc. Eur. Conf. Computer Vision (ECCV)}, 2026.

\bibitem{radford2021clip}
A. Radford et al., ``Learning Transferable Visual Models from Natural
Language Supervision,'' in \emph{Proc. ICML}, pp. 8748--8763, 2021.

\bibitem{liu2024groundingdino}
S. Liu et al., ``Grounding DINO: Marrying DINO with Grounded Pre-Training for
Open-Set Object Detection,'' in \emph{Proc. ECCV}, pp. 38--55, 2024.

\bibitem{kirillov2023sam}
A. Kirillov et al., ``Segment Anything,'' in \emph{Proc. ICCV},
pp. 4015--4026, 2023.

\bibitem{sutton1999options}
R. S. Sutton, D. Precup, and S. Singh, ``Between MDPs and Semi-MDPs: A
Framework for Temporal Abstraction in Reinforcement Learning,''
\emph{Artificial Intelligence}, vol. 112, no. 1--2, pp. 181--211, 1999.

\end{thebibliography}
\endgroup
\end{document}